\documentclass[11pt,a4paper]{article}
\usepackage[hyperref]{emnlp-ijcnlp-2019}
\usepackage{times}
\usepackage{latexsym}

\usepackage{url}

\usepackage{amsmath}
\usepackage{amssymb}
\usepackage{mathtools}
\usepackage{accents}
\usepackage{enumerate}
\usepackage{multirow}

\usepackage{graphicx}
\usepackage{subcaption}
\usepackage{mwe}

\usepackage{booktabs}
\usepackage{graphicx}
\usepackage{graphics}
\usepackage[ruled,linesnumbered,boxed,lined]{algorithm2e}
\usepackage{makecell}

\aclfinalcopy 

\makeatletter
\renewcommand*{\@fnsymbol}[1]{\ensuremath{\ifcase#1\or \mathsection\or \dagger\or \ddagger\or
    \mathsection\or \mathparagraph\or \|\or **\or \dagger\dagger
    \or \ddagger\ddagger \else\@ctrerr\fi}}
\makeatother

\title{The Daunting Task of Real-World Textual Style Transfer Auto-Evaluation}

\author{Richard Yuanzhe Pang \thanks{~\ Work done at the University of Chicago.} \\
  New York University, New York, NY 10011, USA\\
  {\tt yzpang@nyu.edu}}

\date{}

\begin{document}
\maketitle
\begin{abstract}
 The difficulty of textual style transfer lies in the lack of parallel corpora. Numerous advances have been proposed for the unsupervised generation. However, significant problems remain with the auto-evaluation of style transfer tasks. Based on the summary of \citet{pang2018learning} and \citet{mir2019evaluating}, style transfer evaluations rely on three criteria: style accuracy of transferred sentences, content similarity between original and transferred sentences, and fluency of transferred sentences. We elucidate the problematic current state of style transfer research. Given that current tasks do not represent real use cases of style transfer, current auto-evaluation approach is flawed. This discussion aims to bring researchers to think about the future of style transfer and style transfer evaluation research.
\end{abstract}

\section{Introduction}

There are numerous recent works on textual style transfer, the task of changing the style of an input sentence while preserving the content \citep{hu-1,shen-1,fu-1}. One factor that makes textual transfer difficult is the lack of parallel corpora. 
There are abundant advances on developing methods that do not require parallel corpora \citep{simple-transfer,zhang2018learning,logeswaran2018content,yang2018unsupervised}, but significant issues remain with automatic evaluation metrics.
Researchers started by using post-transfer style classification accuracy as the only automatic metric \citep{hu-1,shen-1}. Researchers have then realized the importance of targeting content preservation and fluency in style transfer models, and they have developed corresponding metrics, starting from \citet{fu-1} and \citet{authorship}. 
\citet{pang2018learning} and \citet{mir2019evaluating} have summarized the three evaluation aspects (style accuracy of transferred sentences, content preservation between original and transferred sentences, fluency of transferred sentences) and developed metrics that are well-correlated with human judgments. However, given that current tasks do not represent real use cases of style transfer (Section~\ref{sec:problem-tasks}), we discuss the potential problems of existing metrics when facing real-world style transfer tasks (Section~\ref{sec:content-style}). 
Moreover, \citet{fu-1} and \citet{pang2018learning} have shown that if we obtain different models at different intermediate points in the same training instance, we will get different tradeoffs of style accuracy, content preservation, and fluency. Therefore, more discussions on tradeoff and metric aggregation are needed (Section~\ref{sec:tradeoff}), for better model comparison and selection.

\subsection{Background: Evaluation based on Human-Written ``Gold Standards''}

First, we show that one intuitive way of evaluating style transfer is inadequate: computing BLEU scores \citep{papineni2002bleu} between generated/transferred outputs and human-written gold-standard outputs. In fact, \citet{simple-transfer} crowdsourced 1000 Yelp human-written references as test data (500 positive-sentiment sentences transferred from negative sentiment, and 500 negative-sentiment sentences transferred from positive sentiment). 
From Table~\ref{table:human-500}, we see the striking phenomenon that \emph{untransferred} sentences, compared to transferred sentences generated by the best-performing models, have the highest BLEU score\footnote{We use the \texttt{multi-bleu.perl} script to compute BLEU.} by a large margin. 

\begin{table}[h!]
\centering
\begin{small}
{
\renewcommand{\arraystretch}{0.84}
\begin{tabular}{lcc}
\toprule
Model & BLEU & Accuracy \\
\midrule
\citet{shen-1}\\
\quad CAE$^\dagger$ & 4.9 & 0.818\\
\quad CAE & 6.8 & 0.765\\
\specialrule{.2pt}{1pt}{1pt}
\citet{fu-1}\\
\quad Multi-decoder & 7.6 & 0.792\\
\quad Multi-decoder & 11.2 & 0.525\\
\quad Style embedding & 15.4 & 0.095\\
\specialrule{.2pt}{1pt}{1pt}
\multicolumn{3}{l}{\citet{simple-transfer}}\\
\quad Template & 18.0 & 0.867\\
\quad Delete/Retrieve & 12.6 & 0.909\\
\specialrule{.2pt}{1pt}{1pt}
\multicolumn{3}{l}{\citet{yang2018unsupervised}}\\
\quad LM & 13.4 & 0.854 \\
\quad LM + classifier & 22.3 & 0.900 \\
\specialrule{.2pt}{1pt}{1pt}
\citet{pang2018learning}\\
\quad CAE+losses (model 6) & 22.5 & 0.843\\
\quad CAE+losses (model 6) & 16.3 & 0.897\\
\midrule
Untransferred & \textbf{31.4} & 0.024\\
\bottomrule
\end{tabular}%
}
\end{small}
\caption{Results on Yelp ``style'' (sentiment) transfer. BLEU is between 1000 transferred sentences and 
human references, and accuracy is restricted to the same 1000 sentences. Accuracy: post-transfer style classification accuracy (by a classifier pretrained on the two corpora). CAE$^\dagger$: cross-aligned autoencoder as in \citet{shen-1}. BLEU scores reported for \citet{simple-transfer} are copied from evaluations by \citet{yang2018unsupervised}. \textbf{Note} that if a model name appears twice, the models are from different stopping points during training.
}\label{table:human-500}
\end{table}

This phenomenon either suggests that prior work for this task has not surpassed the baseline of copying the input sentence, or suggests that BLEU is not a good style transfer metric by itself (as it varies by transfer accuracy, as shown in the table). However, it may be a good metric on content preservation, one particular aspect of style transfer evaluation. In fact, \citet{simple-transfer} used BLEU to measure content preservation.

Obtaining human references is costly, and using human references may only solve one aspect of evaluation (i.e., content preservation). We thus complement this aspect and reduce cost by focusing our discussion on automatic evaluation metrics that do not require a large number of references. 

\subsection{Background: Existing \textit{Auto}-Evaluation Metrics}

Researchers have agreed on the following three aspects to evaluate style transfer \citep{mir2019evaluating,pang2018learning}. 

\paragraph{Style accuracy.}

 Style accuracy is the percentage of sentences transferred onto the correct/target style. Automatic evaluation of post-transfer style classification accuracy is computed by using a classifier pretrained on the original corpora. 
Initially, this was the only auto-evaluation approach used in the style transfer works \citep{hu-1,shen-1}.

\paragraph{Content similarity.}

Researchers have realized that when the accuracy is large, the content of the transferred sentence does not necessarily correspond to the content of the original sentence. In particular, \citet{pang2018learning} computed sentence-level content similarity by first averaging GloVe word embeddings \citep{glove} weighted by $\mathrm{idf}$ scores and by computing cosine similarity between the embedding of original sentence and the embedding of the transferred sentence. Next, they averaged the cosine similarities over all original-transferred sentence pairs. The metric has high correlation with human judgments. \citet{mir2019evaluating} first removed style words from the original sentence and the transferred sentence using a style lexicon, and then replaced those words with a $\langle \texttt{customstyle} \rangle$ placeholder. Next, \citet{mir2019evaluating} used METEOR \citep{denkowski-1} and Earth Mover's Distance \citep{pele2009fast} to compute the content similarity. Other works have used similar approaches \citep{cycle-reinforce,simple-transfer,back-translation}, mostly involving BLEU and METEOR \citep{papineni2002bleu,denkowski-1}.

\paragraph{Fluency.}

Researchers realized that style accuracy and content similarity do not guarantee a natural or fluent sentence. \citet{pang2018learning} trained a language model on the concatenation of the original two corpora (of two styles), and used perplexity of resulting transferred sentence to measure fluency. \citet{mir2019evaluating} named the metric ``naturalness,'' and they followed the similar logic with one critical difference. They trained a language model on target style to measure perplexity of transferred sentences. \citet{santos2018fighting} and \citet{yang2018unsupervised} also used perplexity as a measure for naturalness or fluency.

\section{Problem 1: Style Transfer Tasks}\label{sec:problem-tasks}

Before diving into problems of unsupervised auto-evaluation metrics, we first discuss the style transfer tasks in relevant research. The big idea is that we need to move forward from the current operational definition of style, to the real-world and useful definition of style, to be explained below. This transition will create problems for existing style transfer metrics. 

\clearpage

\paragraph{What are the practical use cases of style transfer?}

Here are some possibilities. 
\begin{enumerate}[(i)]
    \item Writing assistance and dialogue \citep{heidorn2000intelligent,ritter2011data}. For example, it is helpful to have programs that transfer a formal sentence to an informal sentence \citep{formality}. It is helpful to have programs that make emails more polite \citep{politeness}.
    
    \item Author obfuscation and anonymity \citep{authorship,gender} so that authors can stay relatively anonymous in, for example, heated political discussions. 
    
    \item For artistic purposes: As an example, we may transfer a modern article to old literature styles. 
    
    \item Adjusting reading difficulty in education \citep{campbell1987adapted}: Programs may be helpful in generating passages of the same content, but of different difficulty levels appropriate to different age groups. 
    
    \item Data augmentation to fix dataset bias \citep{anonymous2020learning}: In sentiment classification using the IMDb movie review dataset \citep{Maas:2011:LWV:2002472.2002491}, for example, the appearance of the word ``romantic'' is highly correlated with positive sentiment, and the appearance of the word ``horror'' is highly correlated with negative sentiment. \citet{anonymous2020learning} thus asked workers to write sentences (where words like ``romantic'' and ``horror'' stay unchanged) with flipped sentiment to reduce spurious correlations. This counterfactual data augmentation approach may also be used to address social bias issues in NLP such as gender, race, and nationality \citep{zhao-etal-2017-men,kiritchenko-mohammad-2018-examining,ws-2019-gender}. Style transfer is a good way to replace most of all of the expensive crowdsourcing procedure. This direction is in line with current NLP community's interest in bias and fairness. 
\end{enumerate}

\paragraph{What do the collected datasets (from the above use cases) look like?}

The two datasets may have very different vocabularies, \textit{and} it is hard to train a classifier to differentiate style-related words from content-related words. As elaborated in Section~\ref{sec:content-style}, certain words need to stay constant despite the fact that the two corpora have drastically different vocabularies. A quick example is that in case (v) above, words like ``romantic'' need to stay unchanged, even if ``romantic'' may not appear in the negative-style vocabulary often. 

Here is another example. In the task of transferring Dickens' style literature to modern style literature but keeping the content \citep{pang2018learning} or in similar literature-related tasks \citep{kabbara2016stylistic,xu2017shakespeare}, the former may contain words like ``English farm'', ``horses'', etc; the latter may contain words like ``vampire'', ``pop music.'' However, these words should stay the same, as they are content-related but not style-related. On the other hand, Dickens' literature may contain words like ``devil-may-care'' and ``flummox'' numerous times, but these words \textit{are} style related and should be changed. Compared to the Yelp sentiment datasets, it is very difficult to automatically differentiate content-related words from style-related words in the Literature dataset. Similar situations may occur frequently in author obfuscation and other practical applications.

\paragraph{Current research focuses on the \textit{operational} definition of style. Those tasks as well as the Yelp sentiment transfer does NOT represent style transfer.}

Therefore, according to the previous paragraph, Yelp sentiment transfer is very idealized, as we can use a simple classifier to classify which words are content-related and which words are style-related. Therefore, changing a word can often change the style (sentiment in this case) successfully. However, to make style transfer useful, we need to go beyond the Yelp sentiment task which most researches focus on. 

In fact, if we generalize the phenomenon, we would find that the current research mostly deals with an \textbf{operational} definition of style where the corpus-specific content words are changed. In the Dickens vs. modern literature example, if the sentence contains the word ``Oliver,'' then it is most likely Dickens style (according to the operational definition), because the word ``Oliver'' has appeared so many times in the novel \textit{Oliver Twist} but the word may have rarely appeared in the modern literature corpus. However, this is not the practical or useful definition of style. 

The vast majority of datasets and use cases are not as idealized as the Yelp dataset. We need to recognize the \textbf{real-world} definition of style (e.g., keeping ``Oliver'' as it is in style transfer), so that style transfer research can show promise of being integrated to application interfaces. This creates problems for the existing automatic evaluation metrics.

\section{Problem 2: The Issue of Metrics}\label{sec:content-style}

\subsection{Content Similarity}\label{sec:problem-sim}

In the task of author obfuscation or writing style transfer, the idea of content similarity becomes rather complicated. In the task of Literature style transfer, what are the style keywords? Take the example where the two unparalleled corpora are Dickens-written sentences and modern literature sentences. 
Consider the following sentence: \textit{Oliver deemed the gathering in York a great success.}
The expected transfer (if we train human annotators/specialists to transfer it) from the Dickens style to the modern literature style should be similar to ``Oliver thought the gathering was successful'' (which is the \textbf{real-world} style transfer). However, the most likely transfer (if we use simple autoencoder framework directly) will be ``Karl enjoyed the party in LA'' (which is the \textbf{operational} style transfer). Consider the following types of words:
\begin{itemize}
    \item Corpus-specific content proper nouns: Names may be different in the transferred sentences, as names in two corpora are different. Similarly for locations, organizations, etc. To transfer correctly, a simple baseline could be using a NER labeller. We can replace words with the corresponding labels, and after transferring the sentence (where some words are represented by labels), we can replace the labels with the original words. In short, these proper nouns need to be consistent. 
    \item Other corpus-specific content words: ``English farms'' should be transferred to ``English farms'' instead of ``baseball fields''; ``horses'' should be transferred to ``horses'' instead of ``vampires.'' In this case, the human-expected rules do not correspond with the machine-identified differences between two corpora. When evaluating, these words are not style keywords, and we should use semantic similarity to make sure that the words stay consistent. 
    \item Style words: ``Deemed'' and ``gathering'' may belong to the Dickens style. They should be changed. 
\end{itemize}
\citet{mir2019evaluating} removed and masked the style keywords by using a classifier. In this case, all of the aforementioned itemized types of words will be masked, and content similarity evaluation will fail. 

We can address this problem by manually creating the list of style keywords, or by retrieving the style keywords by relying on outside knowledge. Another possibility is to keep the words as they are, without removing and masking the style keywords, as the style keywords are likely the minority.

\subsection{Fluency and Style Accuracy}

\citet{pang2018learning} and \citet{mir2019evaluating} both used perplexity. However, one issue is that lower perplexity may reflect unnatural sentences with common words. We can punish abnormally small perplexity as in Section~\ref{sec:solution-4}. Moreover, fluency and style accuracy may have similar problem with Section~\ref{sec:problem-sim}. Perplexity will be large for sentences of the same content but different styles, if the content-words have appeared only rarely in the target corpus. Accuracy has a similar problem. 

Therefore, to address this problem, we can mask out corpus-specific \textit{content} words, before pretraining the language model to evaluate fluency and before pretraining the classifier to evaluate accuracy.

\section{Problem 3: Trade-off and Aggregation of Scores}\label{sec:tradeoff}

Once we have three numbers: style accuracy, content similarity, and fluency, how do practitioners decide which combination to select? According to \citet{pang2018learning} and \citet{mir2019evaluating}, style accuracy is inversely correlated to content similarity, fluency is inversely correlated with content similarity, and fluency is inversely correlated with style accuracy. So how do practitioners determine the degree of trade-off for selection? 

It is often useful to summarize multiple metrics into one number, for ease of tuning and model selection. 
One natural approach is to use aggregation. Suppose we use $A$, $B$, $C$ to represent style accuracy, content similarity, and fluency, respectively. Note that different papers may have different variations of defining $A$, $B$, and $C$. \citet{cycle-reinforce} simply took the geometric mean of $A$ and $B$. However, this choice is arbitrary. In the style transfer models using different datasets, each of $A$, $B$, $C$ corresponds to different range, minimum, and maximum.\footnote{That is, $A$ may fluctuate between 0.4 and 0.6 for models for dataset 1, but $A$ may fluctuate between 0.8 to 0.9 for models for dataset 2. The method of geometric mean does not hold.} Geometric mean is designed so that same percentage change results in same effects of geometric mean. But the percentage change ceases to be meaningful in our case.

\subsection{Potential Solutions for Aggregation}\label{sec:solution-4}

If we still decide to design an aggregation method based on geometric mean, one possible simple remedy similar to \citet{pang2018learning} is to learn a threshold $t_1$, such that $A-t_1$ represents a similar percentage change across many datasets. We define that for sentence $s$,
\begin{align} 
& G_{t_1,t_2,t_3,t_4}(s) = \big( [ A(s) - t_1]_+ \cdot [B(s) - t_2]_+ \cdot \nonumber \\
& \min \{ [t_3-C(s)]_+, [C(s)-t_4]_+ \} \big)^{\frac{1}{3}} \label{eqn:gm}
\end{align}
where $t_1,t_2,t_3,t_4$ are the parameters to be learned as described later. Note that the metric is also designed to punish abnormally small perplexity, as discussed previously. 

One question arises: Is a universal $G$ necessary or helpful (i.e., do we need $G$ that work across many datasets)? The current research strives for a universal metric that work across datasets. If we also strive to do so, we obtain the following result. 

\paragraph{If we need a universal evaluator that works across many datasets.} 

We can randomly sample a few hundred pairs of \textit{transferred} sentences from a range of style transfer outputs (from different models---good ones and bad ones) from a range of style transfer tasks, and ask annotators which of the two transferred sentences is better.\footnote{For each annotation, annotators will be given an original sentence, model-1-transferred sentence, model-2-transferred sentence, and they will be asked to judge which transferred sentence is better if they take all three evaluation aspects into account (style accuracy, content similarity, and fluency).} (Note that the two transferred sentence correspond to the same original sentence). 

We denote a pair of sentences by $(y^+, y^-)$ where $y^+$ is preferred by the annotator. We train the parameters $\mathbf{t}$
using the loss $L(\mathbf{t}) = \max (0, -G_\mathbf{t}(y^+)+G_\mathbf{t}(y^-)+\delta)$
where $\mathbf{t}=\{t_1,t_2,t_3,t_4\}$ and $\delta=1$ as commonly used margin.\footnote{As an example, trained on Yelp dataset and Dickens-modern Literature dataset only, we obtained $t_1=63, t_2=71, t_3=97, t_4=-37$ following the metrics of \citet{pang2018learning}. Please note that this is an extended abstract, so we do not conduct detailed evaluations. To further the quality of the metrics, we propose adding more pairs of transferred sentences from other style transfer tasks to train the parameters $t_1,t_2,t_3,t_4$.}

To even make the metric $G$ more convincing, we may design more complicated functions $G=f(A,B,C)$. Here is a possibility: $G_{\mathbf{t},\boldsymbol{\alpha}}(s) = \big( ([ A(s) - t_1]_+)^{\alpha_1} \cdot ([B(s) - t_2]_+)^{\alpha_2} \cdot \min \{ ([t_3-C(s)]_+)^{\alpha_3}, ([C(s)-t_4]_+)^{\alpha_4} \} \big)^{\frac{1}{3}}$. We can also design $f$ to be a very small neural network (with nonlinear activation), especially if we have lots of annotations. 
We can provide a set of possible function forms $f_1,f_2,\dots,f_p$, and we can train parameters for each individual $f_i$ and select the best $f_i$. We can estimate the quality of $f_i$ by computing the percentage of machine preferences (``which transferred sentence in a pair is better'' according to $G$-scores) that match the human preferences (``which transferred sentence in a pair is better'' according to human judgment).

\paragraph{If we do not need a universal evaluator.}

Then we can repeat the above procedure by only sampling pairs of transferred sentences from the dataset of interest. We suggest this approach, as it will be more accurate for the particular task.

\section{Conclusion}

We discussed existing auto-evaluation metrics for style transfer with non-parallel corpora. We also emphasized that we need to move on from operational style transfer and pay more attention to the real-world style transfer research, so that we can put style transfer systems into practical applications. This shift will create problems for style transfer evaluation metrics. Finally, for ease of model selection of comparison, we discussed possible ways of aggregating the metrics. We hope that this discussion will accelerate the research in \textit{real-world} style transfer.

\section*{Acknowledgements}

The author would like to thank He He and Kevin Gimpel for helpful discussions.

\bibliography{emnlp-ijcnlp-2019}
\bibliographystyle{acl_natbib}

\end{document}